# Décomposition et analyse de tracés EMG pour aider au diagnostic des maladies neuromusculaires

*Arthur Bureau[2], Jean-Maxime Le Carpentier[1], Eric Le Carpentier[1], Yannick Aoustin[1], Yann Péréon[2]*

[1]*Laboratoire des Sciences du Numérique de Nantes, UMR CNRS 6004*
*Ecole Centrale de Nantes & Université de Nantes*
*Nantes, France*

[2]*Laboratoire d'Explorations Fonctionnelles,*
*Centre de Référence Maladies Neuromusculaires Atlantique-Occitanie-Caraïbes,*
*Hôtel-Dieu, Nantes, France*

**Abstract**

L'électromyogramme (EMG) en détection à l'aiguille représente une des étapes de l'électroneuromyogramme (ENMG), examen de réalisation courante en Neurologie. Par l'insertion d'une aiguille dans un muscle et l'étude de la contraction à l'effort, l'EMG fournit des informations extrêmement utiles sur le fonctionnement du système neuromusculaire d'un individu, mais il s'agit d'un examen qui reste complexe à interpréter.

L'objectif de ce travail est de participer à la conception et l'évaluation d'un logiciel permettant une analyse automatisée des tracés EMG de patients suspects de maladies neuromusculaires, orientant le diagnostic vers un processus soit neuropathique, soit myopathique à partir de tracés enregistrés. Le logiciel utilise une méthode de décomposition des signaux selon un modèle Markovien, fondée sur l'analyse de potentiels d'unité motrice obtenus par EMG, puis une classification des tracés. Les tracés de 9 patients ont été ainsi analysés et classés sur la base de l'interprétation clinique du neurologue, permettant d'initier un processus de « machine learning ». Le logiciel sera soumis ensuite à de nouveaux tracés afin de le tester versus un praticien aguerri à l'analyse d'EMG.

*Mots clés : EMG, analyse automatique, traitement de signal, neuropathie, myopathie, maladie neuromusculaire.*

**Introduction**

1.1        *Les maladies neuromusculaires*

Les maladies neuromusculaires touchent entre 40 et 50 000 personnes en France **[1].** Dans la majorité des cas ce sont des maladies rares (incidence < 1/2000), elles peuvent être d'origine génétique (myopathie congénitale, dystrophie musculaire…) ou d'origine acquise (myasthénie grave, neuropathie dysimmunitaire…). Le handicap associé à ces maladies est fréquent, il peut aller de la perte de force et d'utilisation des membres, à la difficulté à avaler, respirer, pouvant amener au décès du patient.
Les neurologues réalisent le diagnostic des maladies neuromusculaires, celui-ci se basant sur les données de l'examen clinique, des analyses biologiques (prélèvement sanguin ou du liquide cérébrospinal), les tests génétiques, la biopsie musculaire, l'imagerie musculaire ou encore l'électroneuromyogramme (ENMG).

Cette étude a pour but l'aide au diagnostic de ces maladies, effectivement, il n'est pas toujours aisé de faire la différence entre un tracé EMG myogène et neurogène, il existe par exemple des tracés dits pseudo-neurogènes ou pseudo-myogènes qui remettent en doute le diagnostic du praticien **[2].**

1.2        *Système neuromusculaire*

La commande motrice volontaire suit un chemin particulier qu'il est important de comprendre si l'on veut étudier ce système.

Un premier neurone part du cortex moteur primaire (gyrus frontal ascendant), se prolonge en direction du tronc cérébral et enfin au niveau de la moelle épinière. A ce stade, on retrouve une synapse entre le premier et seconds neurones moteurs (motoneurones) au niveau des noyaux des nerfs crâniens dans le tronc cérébral ou au niveau de la substance grise de la corne ventrale de la moelle épinière.

Chaque motoneurone envoie à son tour son axone en direction des fibres musculaires qu'il innerve. Très spécifiquement, un axone, correspondant à un motoneurone, vient innerver une seule partie des fibres musculaires composant le muscle. Ces fibres musculaires sont donc regroupées fonctionnellement par leur innervation commune venant du même motoneurone. Cet ensemble « motoneurone/fibres musculaires » est appelé unité motrice (UM). En fonction du muscle étudié, on retrouve plus ou moins d'unités motrices et plus ou moins de fibres musculaires par unité motrice.

Ces motoneurones spinaux sont pourvus de multiples afférences permettant un contrôle fin de la motricité. Ces afférences sont alors traitées, intégrées puis converties en potentiel d'action se prolongeant le long de l'axone en direction des fibres musculaires composant l'UM. La fréquence des potentiels d'action va permettre la régulation de la commande motrice. Une seconde régulation vient du nombre de motoneurone recruté pour générer l'effort musculaire.

Lorsque l'on effectue une contraction musculaire volontaire, la commande consiste en des trains d'impulsions électriques (potentiels d'action) transmis aux muscles par les motoneurones.

Au niveau de la jonction neuromusculaire, les potentiels d'action des motoneurones aboutissent à la transmission synaptique (cholinergique) et génèrent un potentiel d'action musculaire qui se propage sur la membrane du muscle. La contraction des fibres musculaires est ensuite permise par le couplage excitation/contraction.

Cette transmission synaptique explique la correspondance entre potentiel d'action et potentiel musculaire. Pour un potentiel d'action on retrouve un potentiel d'action musculaire ou potentiel d'unité motrice (PUM).

Lors de la contraction, les UM sont activées (« recrutées ») au fur et à mesure, selon la loi des tailles de Henneman, avec des unités motrices dont l'amplitude et la fréquence de décharge augmentent avec l'effort fourni : au début de l'effort, seules les unités motrices les plus excitables, correspondant aux plus petits motoneurones, seront activées et à basse fréquence. Lors d'efforts plus importants, d'autres unités motrices, correspondant aux plus gros motoneurones seront excitées (« recrutement spatial »), avec une fréquence de décharge accrue (« recrutement temporel »). Les caractéristiques des PUM (morphologie, durée, amplitude...) ainsi que le type de recrutement définissent une « signature » du fonctionnement musculaire, variable selon les muscles et modifiée par les pathologies neuromusculaires. Une myopathie peut par exemple modifier certains paramètres de ces PUM, en réduisant leur amplitude et favorisant leur recrutement précoce.

Lors de la contraction, plusieurs unités motrices peuvent être mobilisées et l'on obtient alors un tracé EMG. Celui-ci est donc composé des différents potentiels d'unité motrice qui s'additionnent.

Voici un exemple d'un seul PUM physiologique : **[3]**

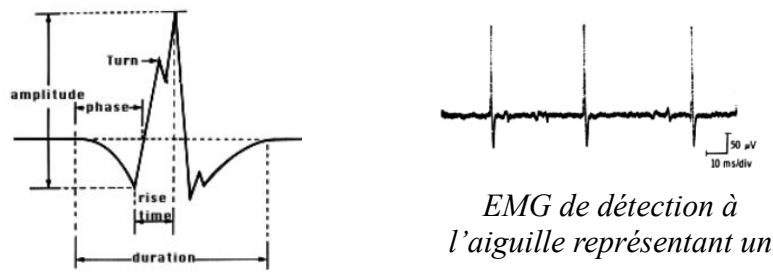

*EMG de détection à l'aiguille représentant un*

Voici à présent un tracé EMG de détection à l'aiguille en contraction moyenne imbriquant plusieurs PUM :

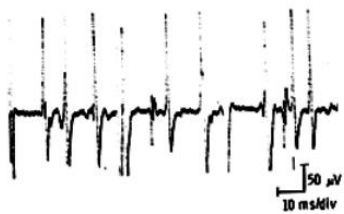

Voici une contraction à effort maximal, dite supramaximale :

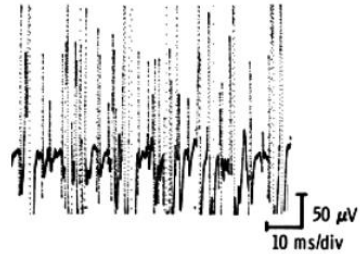

Ce fonctionnement anatomique permet donc d'étudier l'activité électrique des motoneurones et la commande motrice par enregistrement des potentiels musculaires.

L'EMG de détection à l'aiguille permet d'étudier ces UM, et nous donne beaucoup d'information sur le fonctionnement du système neuromusculaire du patient.

### 1.3    *Examen EMG*

Classiquement, le praticien se base sur des paramètres intéressant la totalité du tracé EMG, tels que le recrutement, l'amplitude des PUM, leurs nombres de phase, la richesse du tracé. Ces paramètres sont plutôt qualitatifs et appartiennent à l'interprétation du clinicien. Il existe tout de même des rapports quantitatifs permettant d'orienter le diagnostic, comme le rapport du nombre d'amplitude entre deux « changements de phase » sur le nombre de changements de phase par seconde, à titre d'exemple.

L'examen EMG est dynamique et le praticien va souvent être amené à s'intéresser à plusieurs temps de l'EMG : le muscle au repos, puis augmentation progressive de la contraction volontaire allant jusqu'à la contraction « supramaximale ». C'est globalement sur l'ensemble de l'examen que l'examinateur arrivera à une conclusion diagnostic.

Le but de cette étude est d'améliorer le diagnostic de maladie neuromusculaire : myopathie ou neuropathie, à partir de tracés peu riches en PUM, ce qui diffère beaucoup de la pratique classique de l'examen qui elle essaye d'apprécier un grand nombre de PUM et d'en analyser la résultante.
L'objectif est de recueillir des tracés EMG avec une contraction faible pour n'avoir que quelques PUM, de les isoler et de les analyser. Nous espérons alors retrouver des redondances non physiologiques pouvant être de potentiels marqueurs de ces maladies.

### 1.4    *Pathologies*

Est présenté ici un résumé simplifié des différentes caractéristiques retrouvés chez les patients myopathes et neuropathes, en comparaison aux sujets sains.

Ces tracés EMG ont été obtenus après piqure dans le muscle tibial antérieur de différents patients, au cours d'un effort de contraction de faible intensité, avec recrutement partiel  **[4]** :

Patient de 44 ans sans antécédent de maladie neuromusculaire :

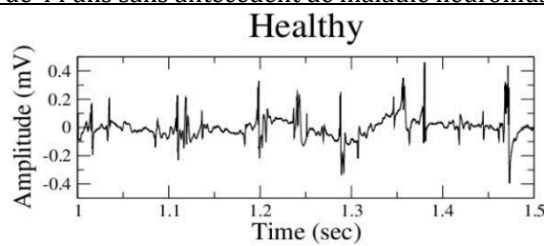

Patient de 62 ans avec des douleurs chroniques dans le dos dues à une radiculopathie L5 :

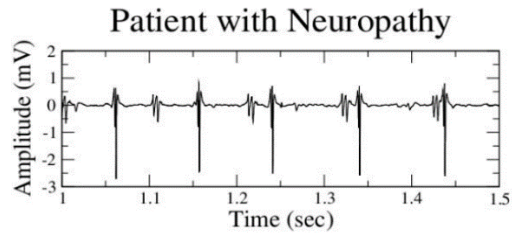

**Neuropathie** : La lésion se situe au niveau du nerf périphérique ou de la racine nerveuse, *cf fig 1*. Moins d'UM sont recrutées mais celles qui le sont on un rating plus élevé (fréquence de décharge plus grande pour les UM encore fonctionnelles). Les fibres musculaires ayant perdu leur motoneurone sont prises en charge par d'autres unités motrices (réinnervation collatérale) augmentant ainsi l'amplitude des PUM avec une contraction synchronisée de plus de fibres qu'habituellement.

De plus, ces phénomènes de réinnervation (post dénervation) engendrent l'apparition « d'unité naissante », c'est-à-dire des PUM de très faible amplitude, de durée allongée, avec plus de changements de phase qu'habituellement, ils sont décrits comme polyphasiques.

Patient de 57 ans atteint de polymyosite (myopathie inflammatoire idiopathique) :

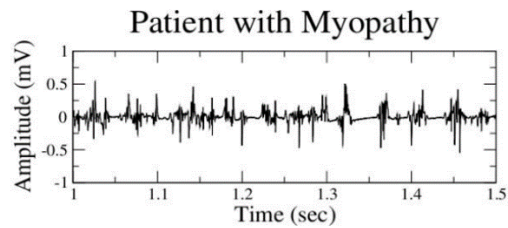

**Myopathie** : Le muscle dégénère et nécrose avec l'évolution de la maladie, *cf fig 1*. On retrouve moins de fibres musculaires activées par UM tout en gardant le nombre total d'UM. Les PUM sont donc de faible amplitude et de durée plus brève. Le recrutement est aussi anormalement élevé pour prévenir le manque d'efficacité musculaire.

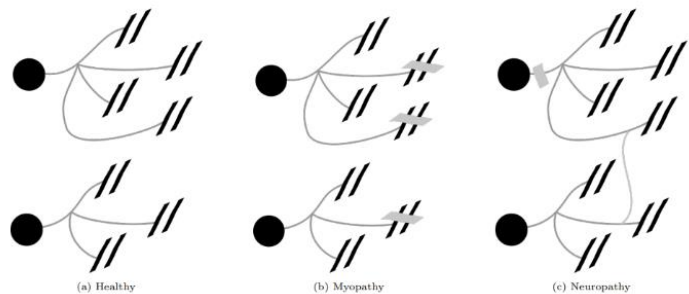

*Fig 1 : Evolution des connexions entre motoneurones et fibres musculaires en fonction des pathologies* **[5]**

Il est intéressant d'ajouter que les nerfs ou muscles atteints dans les neuropathies et myopathies sont généralement de localisation différente.

Les atteintes neurogènes sont à prédominance distales tandis que les atteintes myogènes sont à prédominance proximale (muscles des ceintures). Cela à son importance comme nous le montrera la suite de l'étude, car pour obtenir des tracés EMG spécifiques de la pathologie recherchée, il est préférable d'étudier le muscle atteint ou le muscle lié au nerf atteint.

**Matériel et Méthode**

*3.1 Patients*

Les enregistrements EMG étudiés avaient été réalisés lors de l'ENMG chez des patients atteints de maladies neuromusculaires dans le cadre du soin courant. 9 patients ont été inclus (ratio H/F : 7/2). Il s'agissait de 4 neuropathies (Neuropathie axonale sensitive, Neuropathie axonale sévère diffuse sensitivo-motrice) et 2 myopathies (Tête tombante sans étiologie retrouvée, déficit moteur du quadriceps).

*3.2 EMG de détection*

L'EMG était réalisé au cours de l'examen à l'aide d'une aiguille concentrique standard sur un appareil Natus Keypoint G4. Les signaux étaient filtrés avec une bande passante de 2000 Hz à 4000 kHz, avec une vitesse de défilement de 50 ms/division. Les tracés EMG, d'une durée moyenne de 21,97 secondes, ont été enregistrés sous forme de vidéos numérisées, avec une exportation sous forme exploitable pour le traitement de signal.

*3.3 Matériel*

L'étude a nécessité l'enregistrement d'ENMG intramusculaire permettant une étude électrophysiologique du muscle en obtenant quelques PUM (1 à 10) par tracé.
Cet examen intrusif est encore indispensable pour pouvoir étudier précisément le système neuromusculaire à petite échelle. Effectivement, il existe aussi des ENMG de surface avec des électrodes de surface mais celles-ci ne sont pas assez précises pour distinguer les différentes unités motrices.

Matériel nécessaire à l'examen ENMG :
- Electrode de détection : aiguille concentrique MYOLINE, longueur 30mm, diamètre 0,35mm.
- Câble porte aiguille, 5 pôles DIN, prise mâle 270.
- Câble de terre : câble pince crocodile avec connecteurs 1,5mm TP.
- Appareil EMG : Dantec Keypoint G4

Après avoir obtenus les tracés des différents patients et volontaires sains, il a fallu décomposer les tracés, isoler les potentiels d'unité motrice et enfin les analyser pour ensuite les classer. Ce travail a été possible grâce à une étroite collaboration avec le Laboratoire des Sciences du Numérique de l'Ecole Centrale Nantes. Jean-Maxime Le Carpentier, étudiant en Master sous la direction de Yannick Aoustin et Eric Le Carpentier, s'occupait de cette partie mathématique et informatique.

Logiciel de décomposition et méthodes de classification :

- EMGLAB
- Matlab
- Algorithme d'auto-décomposition

- LDA : Latent Dirichlet Association
- SVM : Support Vector Machine
- Bagging Trees

*3.2 Méthode*

*3.2.1 Obtention des tracés*

La première étape de l'étude était de colliger des données EMG. Pour ce faire, le professeur Yann Péréon m'a permis de l'accompagner en consultation au CHU de Nantes. Chaque patient venait pour un motif de consultation particulier dans le champ des maladies neuromusculaires.
Lorsque la consultation nécessitait un EMG, celui-ci était stocké sur l'appareil puis exporté pour l'étude. Les tracés devaient être franchement myogènes ou franchement neurogènes (par opposition à des cas moins typiques avec des tracés dits pseudo-neurogènes ou pseudo-myogènes). De plus les tracés ne devaient pas être trop riches en PUM pour permettre de bien les individualiser lors de l'analyse post hoc. En pratique nous demandions une contraction faible et essayions d'avoir entre 1 et 10 PUM pendant une dizaine de secondes.

Effectivement une des clefs de la bonne analyse des signaux par les logiciels est la constance du signal.

La constance des signaux étant un élément important pour l'analyse de ces tracés, le mieux aurait été de piquer dans le même muscle pour chaque patient (effectivement chaque muscle a sa « signature » EMG), avec des contractions similaires et avec la même force musculaire générée.

Nous avons alors réfléchi à un protocole de base pour chaque patient : « Une contraction faible, prolongée, de plus de 10 secondes dans le biceps. ». Le biceps à une faible variation interindividuelle et est assez accessible à l'aiguille par rapport à d'autres muscles. De plus il est aisé d'obtenir peu de PUM sur ce muscle dont la contraction est facilement modulable.

Pendant les consultations nous n'avons vu aucun motif concernant le biceps (myopathique ou neuropathique), nous étions contraints de piquer dans les muscles qui intéressaient le patient. Les enregistrements ont été effectués dans le cadre du soin courant, aucun patient n'a eu d'EMG effectué spécifiquement pour notre étude. Le patient était cependant informé du stockage de ses tracés EMG pour utilisation ultérieure, avec son consentement oral, informé et éclairé.

14 tracés neuropathiques ont été enregistrés chez 4 patients dans les muscles vaste latéral et tibial antérieur ; 10 tracés myogènes chez 2 patients dans les muscles splénius du cou et trapèze ; 16 tracés chez 3 volontaires sains dans les muscles vaste latéral et tibial antérieur.

|  | **Muscle** | **Signaux** | **Total** |
|---|---|---|---|
| **Neuropathe** | Vaste latéral | 5 | 14 |
|  | Tibial antérieur | 9 |  |
| **Myopathe** | Splenius du cou | 5 | 10 |
|  | Trapèze | 5 |  |
| **Sain** | Vaste latéral | 7 | 16 |
|  | Tibial antérieur | 9 |  |

Les muscles des volontaires sains ont été choisis en fonction des muscles déjà piqués chez les patients avec atteinte neuropathiques. Cela permet d'avoir une certaine cohérence de signaux entre ces deux populations.

Les muscles piqués chez les patients myopathes sont en accord avec les localisations énoncées plus tôt dans l'étude, à savoir les muscles des ceintures, qui sont plus souvent atteints chez les myopathes.

Voici un exemple des tracés récoltés sur la machine Dantec Keypoint G4 après examen :

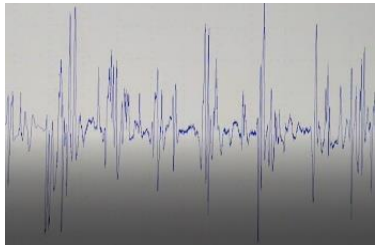

La partie suivante sera un résumé de ma compréhension du principe de décomposition et classification qu'utilisait Jean Maxime Le Carpentier pour l'étude. Elle sera rédigée après une lecture de son mémoire **[5]** qui explique les versants mathématiques bien plus profondément, ainsi qu'après de multiples discussions lors de nos temps de travail. Le but est ici de comprendre le principe général de la décomposition et de toucher du doigt les mécanismes de classification.

*3.2.2 Décomposition des signaux*

La décomposition correspond à l'isolement des potentiels d'unité motrice qui se répètent durant le temps du tracé. Le logiciel va être capable de reconnaître les PUM se répétant dans le temps et va les isoler. Cela permettra par la suite de les analyser un par un.

Après l'obtention des tracés sur la machine *Dantec Keypoint G4,* , ceux-ci ont été exportés dans des fichiers en format « .txt », seul format pouvant être utilisé par les logiciels de décomposition. Le tracé était alors converti en fichier texte avec une fréquence d'échantillonnage de 24 000 Hz et avait pour unité le microvolt. Cet enregistrement comprenait l'entièreté du signal enregistré sur la machine EMG :

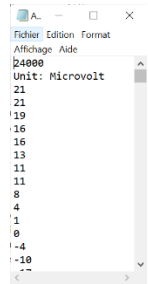

Cet échantillonnage a finalement permis d'obtenir les tracés de la machine EMG sur le logiciel Matlab :

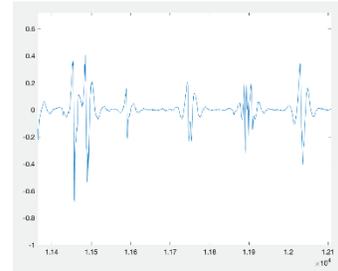

Depuis Matlab, ces tracés étaient ensuite analysés par EMGLAB, c'est réellement cet outil qui permet de décomposer le signal EMG.

Par exemple, sur illustration suivante, le logiciel a identifié au moins 8 PUM différents se répétant le long du tracé EMG, chaque PUM possédant ses propres caractéristiques :

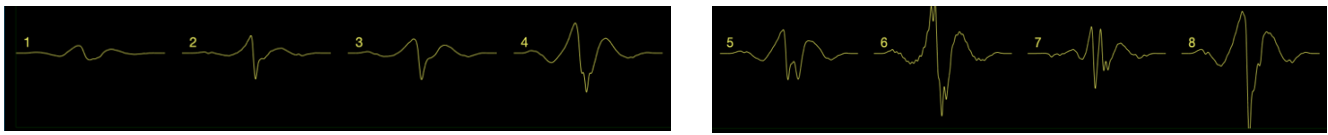

Aux PUM sont attribués des numéros pour pouvoir les différencier. Le logiciel permet aussi d'obtenir un tableau à double entrée avec à gauche le temps auquel le PUM apparaît et à droite le « numéro du PUM » :

```
0.0156    5.0000
0.0168    6.0000
0.0188    3.0000
0.0231    2.0000
0.0316    1.0000
0.0320    4.0000
0.0325    7.0000
0.1096    2.0000
0.1098    1.0000
0.1114    5.0000
0.1117    3.0000
0.1245    4.0000
0.1278    6.0000
0.1719    1.0000
0.1843    2.0000
0.1962    3.0000
0.2038    5.0000
0.2188    4.0000
```

Il est important de préciser que cette décomposition est semi-manuelle, c'est à l'opérateur de choisir quel signal proposé par le logiciel est réellement un PUM ou non. Cela nécessite donc de savoir reconnaître les potentiels après leur échantillonnage et après une transformation sur un autre logiciel. Effectivement les tracés obtenus sur la machine EMG et ceux obtenus sur EMGLAB sont parfois assez distincts visuellement et il n'est pas toujours aisé de reconnaître un vrai PUM ni de savoir éliminer les mauvais. De plus cette méthode est assez chronophage.

En parallèle, une méthode de décomposition automatique spécifiquement élaborée au sein du Laboratoire des Sciences du Numérique de l'Ecole Centrale a été utilisée.
Créé par Tianyi Yu, un doctorant de l'école Centrale Nantes **[7], [9],** cet algorithme fût adapté pour chaque signal par Jean Maxime en modifiant les paramètres. L'algorithme a été spécialement créé dans le but de décomposer des signaux EMG.
Nous obtenons alors des tracés décomposés avec des PUM différenciés. Ces PUM ont des caractéristiques qui peuvent être extraites et cela va nous permettre de les classer.

*3.2.3 Classification*

La classification a pour but de classer les différents signaux EMG à travers l'étude de tous les PUM qui composent les signaux. Elle ne classe pas seulement chaque PUM indépendamment les uns des autres mais bien un ensemble de PUM correspondant à un tracé EMG. Cela a pour objectif de classer directement un signal EMG en myopathique, neuropathique ou en sain.

La classification va se baser sur les paramètres que l'on peut extraire de chaque unité motrice issue de la décomposition du signal. Le principe général est de sélectionner les principales caractéristiques des PUM (amplitude, durée, nombre de changements de phase, firing rate…), et de leur attribuer des vecteurs (valeur max ; valeur min ; moyenne). Ces différents vecteurs (un vecteur pour un PUM) vont constituer une caractérisation de chaque signal.

En totalité, 8 caractéristiques ont été utilisées pour classer les PUM, 7 d'entre eux ayant 3 modalités (min ; max ; moy), nous donnant alors 22 paramètres. Le logiciel Matlab permet de sélectionner les paramètres les plus pertinents pour la classification et ce sont les 18 suivants qui ont guidés la classification pour la suite de l'étude :

|                | Minimum | Mean | Maximum |
|----------------|---------|------|---------|
| nb MUAPs       |         |      |         |
| firing rate    | x       | x    | x       |
| amplitude      | x       | x    | x       |
| nb phases      |         | x    | x       |
| energy         |         | x    | x       |
| nb turns       |         | x    | x       |
| duration       | x       | x    | x       |
| time spreading | x       | x    | x       |

C'est ainsi que Jean Maxime a sélectionné 3 méthodes de classification pour les comparer entre elles.

Ces trois méthodes sont :

- **LDA** pour « linear discriminant analysis »
- **SVM** pour « Support Vector Machine »
- **Bagging Trees**

Ces différentes techniques utilisent des procédés mathématiques qui vont bien au-delà de mes compétences. Pour essayer de résumer ce que m'en a expliqué Jean-Maxime, elles permettent, à partir des différents paramètres énoncés précédemment, de scinder les populations de tracés EMG de part et d'autre d'une droite pour la LDA et la SVM.

LDA            SVM

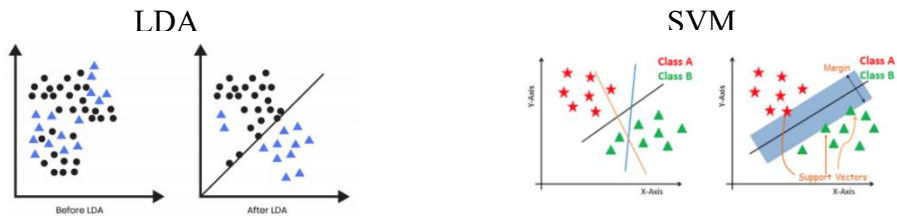

La méthode Bagging Trees est basée sur la création d'arbre de décision à partir de signaux EMG qui servent à « entraîner » le processus. Ensuite le signal test est analysé par ces différents arbres et en ressort une classification.

Bagging trees

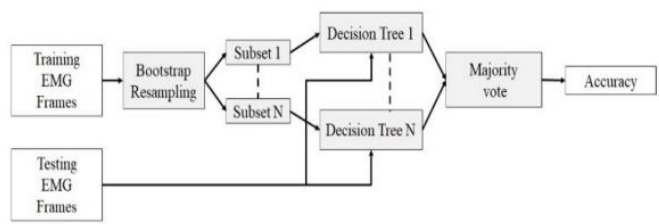

Pour ces méthodes, il est nécessaire d'avoir des signaux qui entraînent le classificateur. Il a été décidé que 30% des signaux (3 myogènes, 4 neurogènes et 5 sains) serviraient de test après entraînement et que 70% des autres signaux obtenus pendant la collecte de donnée (28 signaux) permettraient l'entraînement. Les signaux test ont été tirés au sort.

Voici donc la méthode utilisée pour classer nos signaux EMG en neuropathique, myopathique ou sain.

Pour plus d'explications sur cette méthode, se référer au travail de Jean Maxime Le Carpentier **[6]**.

## 4. Résultats

Sont exposés les résultats des différentes méthodes de classification concernant la base de données créée pour l'étude au CHU de Nantes. Le travail de Jean Maxime Le Carpentier est complémentaire et expose d'autres résultats dans son travail, avec des données (signaux EMG) extérieures à cette étude, dans le but de comparer les 3 méthodes entre elles.

Résultats selon les différentes méthodes :

|     | Accuracy | Time    |
| --- | -------- | ------- |
| LDA | 66.67%   | 4.7ms   |
| SVM | 58.33%   | 35.2ms  |
| BT  | 66.67%   | 213.4ms |

Ce tableau expose dans la colonne de gauche la précision des classificateurs à donner le bon diagnostic par rapport à l'évaluation d'un praticien aguerri, ici le professeur Yann Péréon. Dans la colonne de droite on retrouve le temps qu'à mis le programme à donner sa classification.

Le classificateur utilisant la SVM obtient une précision de 58,33%, les deux autres classificateurs LDA et Bagging Trees obtiennent des précisions de 66,67%.

## 5. Discussion

Durant cette étude, nous avons évalué une méthode d'automatisation de décomposition et de classification d'EMG à partir de tracés de novo de patients, dans le but savoir si cette méthode était pertinente vis à vis de l'aide au diagnostic de maladies neuromusculaires.
L'automatisation de l'analyse d'EMG a commencé dès les années 50 et s'est développée dans les années 80 avec l'évolution des machines de calcul, notamment à Nantes avec le professeur Pierre Guihéneuc, sans succès.

Les résultats de l'étude indiquent que ces méthodes de décomposition et de classification sont encore insuffisantes pour une aide correcte au diagnostic, dans le cadre de l'étude.
Cependant, les résultats sont tout de même encourageants. Il est utile de rappeler que le nombre de signaux pour l'étude ainsi que les conditions d'enregistrement ne sont pas optimales pour ce type de méthode qui nécessite beaucoup de signaux pris dans les mêmes conditions. La précision des classificateurs dépend directement des tracés EMG récoltés car ce sont eux qui permettent d'entraîner les programmes. On peut alors imaginer qu'avec plus de tracés et avec des signaux EMG enregistrés de façon plus sélective, la précision augmenterait.

Dans son étude, Jean Maxime Le Carpentier a d'ailleurs testé ces méthodes de décomposition puis de classification sur d'autres data que celles de l'étude. Ces autres données correspondent à des signaux EMG franchement neurogène, myogène ou sain, venant du même muscle, prises sur des banques de données pouvant être traitées sur Matlab/EMGLAB.
Avec un nombre bien plus important de tracés et des signaux plus comparables, on retrouve des précisions plus élevées que dans notre étude :

|     | Accuracy | Time    |
| --- | -------- | ------- |
| LDA | 66.67%   | 6.0ms   |
| SVM | 80%      | 22.9ms  |
| BT  | 80%      | 190.2ms |

Une autre limite de l'étude vient directement de l'examen EMG. L'EMG de détection à l'aiguille permet d'enregistrer des potentiels d'unité motrice d'une seule partie du muscle et n'est donc pas exhaustif. Effectivement, l'aiguille est très fine et ne permet d'apprécier que quelques fibres musculaires sur l'ensemble du muscle étudié.

De plus, les muscles étudiés pour chaque patient sont différents, or nous savons que chaque muscle à sa « signature » EMG, cela influence donc les résultats quand il s'agit de comparer différents tracés. La réalisation d'un protocole n'était pas applicable à l'étude car les patients inclus venaient pour un motif de consultation précis et propre, les examens étant réalisés en tant que soins courants. Ce mode d'inclusion nous a permis de recueillir un bon nombre de tracés, mais ce nombre pourrait être encore largement augmenté, ce qui augmenterait l'entraînement des programmes et ainsi, théoriquement, la précision de ces derniers.

Ces différents éléments montrent que des améliorations peuvent être apportées pour obtenir des résultats plus précis concernant la méthode de recueil de données.

Il existe aujourd'hui de nombreuses études qui expérimentent des méthodes d'analyse automatique suivant le même schéma que le nôtre : décomposition puis classification. Cependant ces études permettent de classer des PUM tandis que le but de notre étude est de classer des signaux EMG regroupant plusieurs PUM différents.

D'autres méthodes de décomposition et de classification existent et il est intéressant de voir que les données sont très diverses en fonction des études et modèles utilisés.

Sans rentrer dans le détail technique des classificateurs voici un tableau résumant des résultats de certaines études menées sur le sujet en EMG à l'aiguille visant à classer des PUM [8] :

| Classifieur | Auteurs | Nb patients ou PUM | Précision / Accuracy | Commentaires |
|---|---|---|---|---|
| *ANN BP* | Pattichis & al - 1995 | 44 patients : 16 N<br>14 M<br>14 H | 80-90% | Contraction volontaire biceps, supervised learning |
| *ANN Kohnen SOFM model* | Pattichis & al - 1995 | 44 patients : 16 N<br>14 M<br>14 H | 80-85% | Contraction volontaire biceps, unsupervised learning |
| *ANN SOFM + LVQ* | Bhardwaj & al - 2012 | 11 patients :<br>5 N<br>4 M<br>2 H | 97,6% | |
| *Fuzzy system* | Chauvet & al - 2001 | | 88,4% | |
| *Hybride NFS* | Koçer & al - 2010 | 177 PUM<br>60 N<br>60 N<br>57 H | 90% | (Classificateur avec 3 layer MLP) |
| *SVM* | Kastsis & al - 2006 | 231 PUM | 86,14% | |

*Légende :*
- *Pour les patients / PUM : N = neuropathie ; M = myopathie ; H = Healthy.*
- *ANN = Artificial neural network*
- *BP = Back propagation*
- *SOFM = Self organizing feature map*
- *LVQ = Leaving Vector quantization*
- *NFS = Neurofuzzy system*
- *MLP = Multilayer perception*

Cette aide automatisée au diagnostic serait un réel avantage dans les nombreux cas où les tracés sont ambigus pour les cliniciens. Les tracés pseudo-neurogènes, pseudo-myogènes, ou en désaccord avec le reste du tableau clinique sont nombreux et méritent d'être précisés pour aider au diagnostic mais pourrait aussi servir au suivi des affections neuromusculaires.

A terme, l'objectif serait d'implanter dans la machine à EMG, un logiciel permettant d'analyser en direct les tracés, le praticien pourrait alors avoir un aiguillage presque instantané sur les résultats de l'examen. Le but n'est pas de remplacer le praticien, qui est indispensable à la mise en place de l'examen et son analyse, mais bien de lui donner des outils supplémentaires pour améliorer sa précision diagnostic et permettre au patient d'être mieux pris en charge.

**Conclusion**

Ces résultats nous encouragent à penser qu'une étude plus conséquente, avec plus de signaux et un protocole respecté, pourrait apporter une meilleure précision à la méthode d'analyse utilisée.

Cette aide au diagnostic pourrait avoir un impact non négligeable dans les nombreux cas où les tracés sont difficilement évaluables. L'étude précise et automatisée des PUM aiguillerait le praticien qui pourrait alors avoir un nouvel outil dans sa panoplie de tests diagnostic.

**Remerciements**

**Références**

- L'électromyographie sans douleur volume 1, édition Lavoisier, 2014, Emmanuel Fournier

- Sémiologie EMG élémentaire volume 2, édition Lavoisier, 2013, Emmanuel Fournier

**[1]** https://www.filnemus.fr/les-maladies/domaine-dexpertise-de-la-filiere

[2] Y. Pereon, Neurogène ou myogène, Revue Neurologique 171S 1189 (2015)

[3] https://teleemg.com/manual/the-motor-unit/

[4] EMG studies (By Seward Rutkove, MD, Department of Neurology, Beth Israel Deaconess Medical Center/Harvard Medical School

[5] A.~Gallard, K.~Akhmadeev, E.~Le Carpentier, R. Gross, Y. Péréon, and Y.~Aoustin, ''Automatic Classification of Intramuscular EMG to Recognize Pathologies'' In Abali B., Giogio I. (eds) Developments and Novel Approaches in Biomechanics and Metamaterials Advanced Structured Materials, Vol 132, Springer, Cham, https://doi.org/10.1007/978-3-030-50464-9_3.

[6] Jean-Maxime LE CARPENTIER, Arthur BUREAU, Konstantin AKHMADEEV, Tianyi YU, Yannick AOUSTIN, Eric LE CARPENTIER, Yann PEREON, Recognition, Analysis, Decomposition of EMG Recordings for the diagnosis of neuromuscular diseases – Master's Thesis

[7] Yu, Tianyi. « On-line decomposition of iEMG signals using GPU-implemented Bayesian filtering ». Theses, École centrale de Nantes, 2019. https://tel.archives-ouvertes.fr/tel-02407288.

[8] Yousefi, Jamileh, et Andrew Hamilton-Wright. « Characterizing EMG Data Using Machine-Learning Tools ». *Computers in Biology and Medicine* 51 (1 août 2014): 1-13. https://doi.org/10.1016/j.compbiomed.2014.04.018.

[9] T.~Yu, K.~Akhmadeev, E.~Le Carpentier, Y. Aoustin, R.~Gross, Y.~Pereon, and D.~Farina. Recursive decomposition of electromyographic signals with a varying number of active sources: Bayesian modelling and filtering, IEEE Transactions on Biomedical Engineering, Vol. 67, (2), pages 428-440, 2020.